\documentclass{article}

\usepackage[final]{nips_2016}

\usepackage[utf8]{inputenc} 
\usepackage[T1]{fontenc}    
\usepackage{hyperref}       
\usepackage{url}            
\usepackage{booktabs}       
\usepackage{amsfonts}       
\usepackage{nicefrac}       
\usepackage{microtype}      
\usepackage{dutchcal}

\usepackage{amsmath}
\usepackage{float}
\usepackage{graphicx}
\usepackage{amssymb}

\title{A simple squared-error reformulation for ordinal classification}

\author{
  Christopher Beckham \& Christopher Pal\\
  Polytechnique Montr\'{e}al \& \\
  Montr\'{e}al Institute for Learning Algorithms \\
  Université de Montr\'{e}al \\
  \texttt{firstname.lastname@polymtl.ca} \\
}

\begin{document}

\maketitle

\begin{abstract}
  In this paper, we explore ordinal classification (in the context of deep neural networks) through a simple modification of the squared error loss which not only allows it to not only be sensitive to class ordering, but also allows the possibility of having a discrete probability distribution over the classes. Our formulation is based on the use of a softmax hidden layer, which has received relatively little attention in the literature. We empirically evaluate its performance on the Kaggle diabetic retinopathy dataset, an ordinal and high-resolution dataset and show that it outperforms all of the baselines employed.
\end{abstract}

\section{Introduction}

Ordinal classification is a  form of classification in which there is a natural ordering between the classes to be predicted. This implies that some misclassifications are worse than others, and that the distances between classes be taken into account when training or evaluating a classifier -- this is in contrast to discrete classification, in which there are no misclassifications that are worse than others.

Gutierrez et al. \cite{ordinal_review} proposed and created a taxonomic tree to classify different kinds of ordinal classification techniques, these being: naive techniques; ordinal binary decompositions; and threshold-based approaches. Naive techniques include treating the problem like a regression or employing cost-sensitive classification \cite{conditional_risk}; ordinal binary decompositions involve techniques like training several binary classifiers \cite{simple_ordinal} or training a single classifier that can handle multiple outputs (say, a neural network) \cite{hacky_ord_approach}; and threshold-based approaches include cumulative link models \cite{classical_models} and ensembles, to name a few. In order to limit the scope of our work, we devise a new technique for neural networks that falls under the `ordinal binary decomposition' category, which includes single models with multiple outputs, i.e., neural networks. This technique is appealing as it only involves the training of a single model, as opposed to multiple model approaches where the ordinal classification problem is decomposed into simpler binary problems. However, our technique can also be thought of as a naive approach according to this taxonomy, since the overall technique is based on squared error (i.e., regression), and this assumes that classes are equally distant from each other. In the case of training, we hypothesise that our method outperforms a cross-entropy loss (i.e., the loss employed when treating the problem as a discrete classification) due to the relaxation of the constraint that the probability mass be totally concentrated on the correct class. We believe this can especially be beneficial when there is significant labelling noise. Furthermore, our technique also learns produces probability distributions over the data.

Deep neural networks have enjoyed recent success in image classification and other realms, achieving state-of-the-art results on many benchmark tasks \cite{imagenet, delving, inception}. Consequently, they have countless applications in medical contexts. Furthermore, many problems in medicine are of an ordinal nature (e.g. diseases come in multiple stages), which motivates our work in the exploration of ordinal classification methods in the context of deep neural networks.

\subsection{Architectures for Ordinal Predictions}

There are a number of ways in which an ordinal prediction can be made with a deep network. One way is to simply perform a regression, using a squared error loss, or using cross-entropy.

We examine here an architecture where we use a softmax \emph{hidden layer} which modulates the mean of a Gaussian distribution which is used to compute the loss.  The parameterization is simply a normal distribution with mean given by $\textbf{a}^T\textbf{f}(\textbf{x})$ and variance $\sigma^{2}$ such that 
%
\begin{equation}
p(c|\textbf{x})=
\mathcal{N}(c;\textbf{a}^T\textbf{f}(\textbf{x});\sigma^{2})
\end{equation}
where $\textbf{a} = [0, 1, \dots, k-1]^{T}$, $\textbf{f}(\textbf{x})$ is the probability distribution of $c$ over $\textbf{x}$, and $c$ is a scalar real random variable which is observed in practice as an integer between 0 and $k-1$. In this way, each of the softmax hidden units could be interpreted as a prediction for each ordinal category. 

%
%
%
%
\subsection{Data}

In early 2015, Kaggle\footnote{https://www.kaggle.com/c/diabetic-retinopathy-detection/}---a crowdsourcing platform for predictive modelling---released a large amount of labeled fundus images (i.e., images of the interior surface at the back of the eye) as part of a competition to try and find the best predictive models for diabetic retinopathy detection. The fundus image data are provided in the form of high-resolution images taken under various imaging conditions. Left and right eye images are provided for every patient, and the training and testing sets comprise 35,126 and 53,576 images, respectively. The goal is, given a fundus image of either the left or right eye, to predict its class: no DR (25,810 images), mild DR (2,443 images), moderate DR (5,292 images), severe DR (873 images), or proliferative DR (708 images).

Because we are dealing with ordinal data, an appropriate evaluation metric to use is the quadratic weighted kappa, which, as the name suggests, imposes a penalty that is proportional to the distance between the predicted class and actual class. Interestingly, though it is not the main focus of this paper, we tried to optimise the quadratic weighted kappa directly and also obtained promising results. This can be found in Appendix 1, and we plan to talk about this more in future work.

\section{Methods and Results}

When training neural networks on ordinal data, one of the simplest techniques is to simply treat the problem as a discrete classification and employ the cross-entropy loss:
\begin{equation} \label{eq:x_ent}
-\sum_{i=0}^{k-1}\textbf{y}_{i} \log( \textbf{f(\textbf{x})}_{i} ),
\end{equation}
where $\textbf{f(\textbf{x})}$ is the output layer of the deep net with a softmax activation.

\begin{figure}
    \centering
    \includegraphics[width=0.3\textwidth]{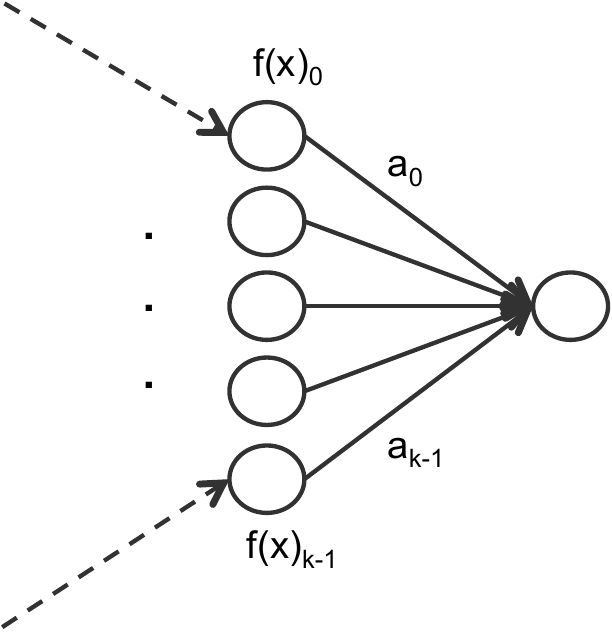}
    \caption{The final layers of an arbitrary neural network. The penultimate layer $f(\textbf{x})$ in this example has $k = 5$ units. The final layer (consisting of one hidden unit) computes $p(c|\textbf{x})$ and is parameterized by weight vector $\textbf{a}$.}
    \label{fig:my_label}
\end{figure}

Because the quadratic weighted kappa can severely penalise certain misclassifications, we can generate `conservative' predictions by computing the `soft' argmax \cite{argmax} of $p(c|\textbf{x})$ by computing $\textbf{a}^{T}\textbf{f(\textbf{x})}$. This trick can also be thought of as tacking on an extra layer at the end of the (final) softmax layer where the weights are fixed, as illustrated in Figure \ref{fig:my_label} (where $\textbf{a}_i = i$). This has the benefit of introducing no extra parameters to be learned for the network. Therefore, in our first experiment, we simply employ the cross-entropy loss, and we will refer to this experiment as \texttt{cross-entropy}. The second experiment, which is our technique, is simply the squared error $(c - \textbf{a}^{T}f(\textbf{x}))^2$, which we will refer to as \texttt{fix 'a'} (where \texttt{fix} refers to the fact that the vector $\textbf{a}$ is fixed).

Note that we compute predictions on the validation set by computing $\textbf{a}^{T}f(\textbf{x})$ and rounding to the nearest whole integer, instead of the argmax of $f(\textbf{x})$ which is typically employed, as we have found this consistently generates more accurate predictions, due to its conservative nature. One downside however is that the output of the network, $\textbf{a}^{T}\textbf{f(\textbf{x})}$, is a continuous prediction rather than a discrete one, and this means some rounding scheme must be employed. As an interesting side note, Domingos \cite{metacost} proposed a simple technique to generate cost-sensitive predictions that are discrete through the minimisation of conditional risk, and we have found that our technique and his almost always generate the same quadratic weighted kappa on our validation set. Because of this, we are comfortable with simply rounding our predictions to the nearest whole integer.

As another baseline, we employ the technique presented in Cheng \cite{hacky_ord_approach}, in which $\textbf{y} \in \{0,1\}^{k-1}$ (as opposed to $\textbf{y} \in \{0, \dots, k-1\}$). In this case, if we have $k$ classes, each $\textbf{y}$ is a binary code where the first class is $[0, \dots, 0]^{T}$, the second class is $[1, \dots, 0]^{T}$, the third class is $[1, 1, \dots, 0]^{T}$ and so on. In this case, $\textbf{f(\textbf{x})}$ instead has the sigmoid nonlinearity applied to it, and the loss becomes the binary cross-entropy:
\begin{equation}
\sum_{i} -\textbf{y}_{i}\log(\textbf{f(\textbf{x})}_{i}) - (1 - \textbf{y}_{i})\log(1 - \textbf{f(\textbf{x})}_{i})
\end{equation}
Suppose we have an output code $\textbf{f(\textbf{x})} \in [0,1]^{k-1}$, where $\textbf{f(\textbf{x})}_{i} = p(c > i | \textbf{x})$, we threshold each code to either 0 or 1 depending on if it is greater than or equal to 0.5. Then, the index of the first zero in the code $j$ is the predicted class. For example, if $\textbf{f(\textbf{x})} = [1,1,0,0]^{T}$ then $j = 2$. However, since each code in the output is independent, it is possible to obtain erroneous codes such as $[1,1,0,1]^{T}$ (i.e., not monotonically decreasing), in which case the predicted class is still $j = 2$. If there are no zeroes in the code, then the prediction is the maximal class $k-1$. We name this experiment \texttt{cheng}.

Note that we could also learn the weights in $\textbf{a}$ and use the squared error loss, in which case the problem becomes a typical regression. This constitutes another baseline (even though there are $k$ extra parameters that can be learned), which we call \texttt{learn 'a'} (since we are learning the vector $\textbf{a}$). For this experiment, we employ the softmax for $f(\textbf{x})$, to be consistent with \texttt{fix 'a'}. Note that when we learn $\textbf{a}$, $\textbf{a}^{T}f(\textbf{x})$ is unbounded, which can be detrimental to learning, so we also run an experiment where we constrain the output by optimising $(c - (k-1)\sigma(\textbf{a}^{T}f(\textbf{x})))^2$, where $\sigma$ is the sigmoid nonlinearity, and we multiply by $k-1$ so that predictions lie between 0 and $k-1$. We call this experiment \texttt{learn 'a' (sigm)}.

We perform two runs of each experiment. For all experiments, we use Nesterov momentum with initial learning rate $\alpha = 0.01$, momentum 0.9, and batch size 128, with the exception of \texttt{fix 'a'}, where $\alpha = 0.1$ initially and decreased to $\alpha = 0.01$ at 61 epochs for the first run, and $\alpha = 0.01$ at 118 epochs for the second run. Finally, all experiments have $\alpha$ decreased to 0.001 after 200 epochs. Data augmentation techniques such as random crops, rotations, horizontal, vertical flips are used. We use a modest ResNet \cite{resnets} architecture for our network.

We use Theano \cite{theano} and Lasagne \cite{lasagne} to run our experiments. The results of these are shown below in Figure \ref{fig:my_label2}.

Illustrated in the left plot is the cross-entropy loss computed on the validation set for the \texttt{cross-entropy} and \texttt{fix 'a'} experiments. Interestingly, we can see that the experiment \texttt{fix 'a'} achieves a lower validation loss than the baseline \texttt{cross-entropy}, even if convergence is slow initially. This is a surprising result, since the loss employed is a squared-error loss and not cross-entropy. (Other experiments are not plotted since evaluating the cross-entropy on $\textbf{f}(\textbf{x})$ is senseless.)

\newpage

\begin{figure}[H]
    \centering
    \includegraphics[width=0.9\textwidth]{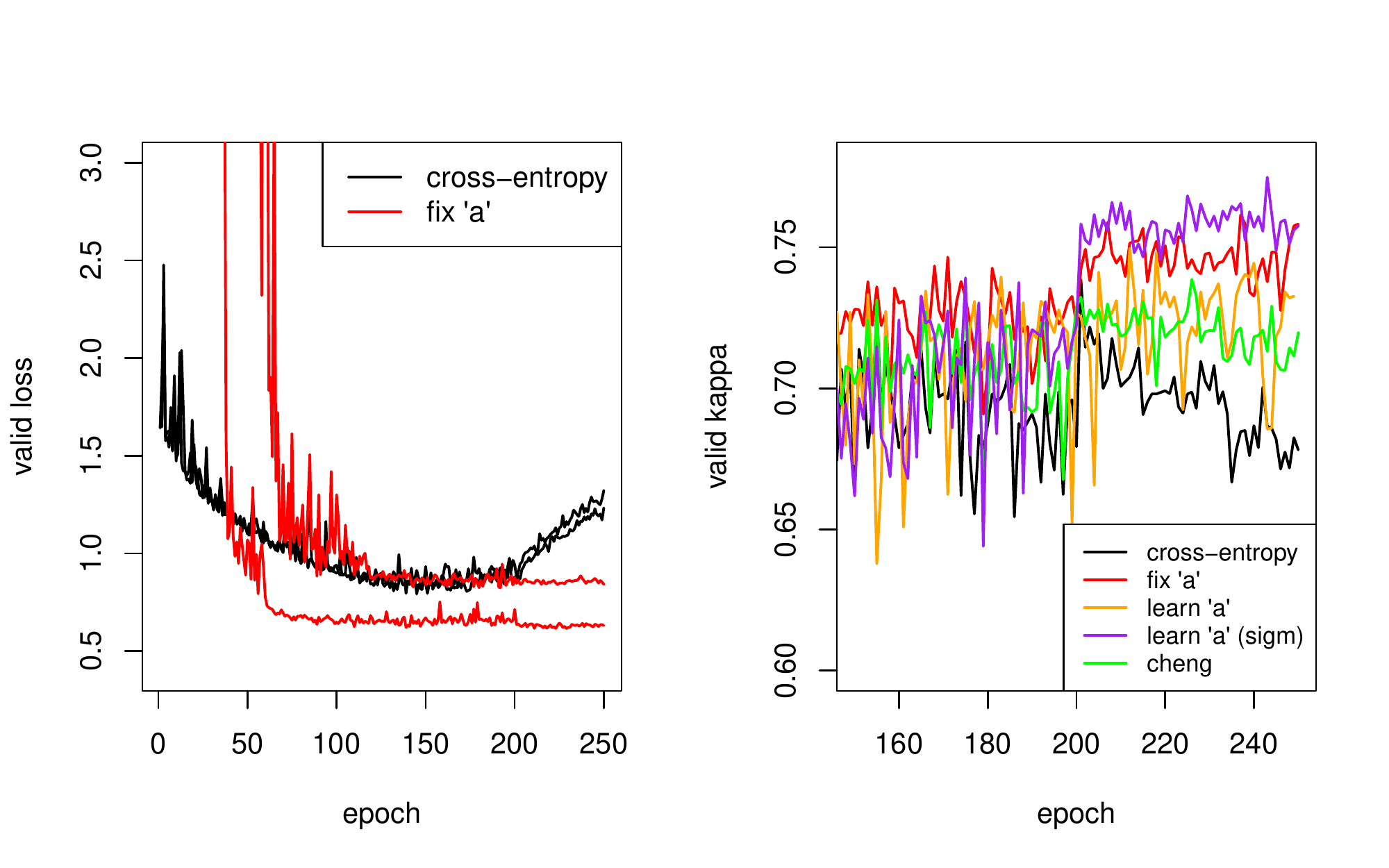}
    \caption{Categorical cross-entropy loss (left) and quadratic weighted kappa (right) for the experiments. (Each experiment on the right is an average of its two constituent runs.)}
    \label{fig:my_label2}
\end{figure}



The right-most plot shows the validation set kappa for all experiments run. We can see that \texttt{fix 'a'} consistently achieves a much higher QWK than \texttt{cross-entropy} and the ordinal encoding scheme proposed by Cheng. However, the technique is outperformed by \texttt{learn 'a' (sigm)}, even though this technique has $k$ extra learnable parameters. Despite this, \texttt{fix 'a'} is unique in the sense that we learn probability distributions over the data, which is not possible under the other experiments.

\section{Conclusion}

In this paper we present a simple ordinal classification technique whose formulation is a Gaussian distribution whose mean is based on a softmax hidden layer. When we use this model in conjunction with a squared error loss, we not only obtain a loss that is sensitive to class ordering (due to the utilization of squared error), we can still obtain a discrete probability distribution over the class through $f(\textbf{x})$. This is in contrast to employing a typical regression, where we let $\textbf{a}$ be a learnable parameter, though $f(\textbf{x})$ (when it has a softmax nonlinearity) is not guaranteed to produce a valid probability distribution over the classes. We have shown empirically that our method produces competitive results with other techniques we have evaluated, with the added benefit of still having a probability distribution over the classes. We note however that our technique shares a common weakness with regression, in the sense that classes are equidistant from each other.

The softmax parameterization is frequently used for the final layer of a neural network. While it is known that softmax hidden layers are possible, it is not common in the literature to use softmax hidden layers. These experiments thus provide some interesting evidence on the utility of softmax hidden layers.

\subsubsection*{Acknowledgments}

We thank Samsung for supporting this research.

\section{Appendix}

In this section we briefly explore the quadratic weighted kappa metric and our attempt at directly optimising it. To our knowledge, we have only seen two attempts in the literature that try to optimise this: Vaughn and Justice \cite{derek} attempt to derive the QWK in a way that makes it easily optimisable (though as of time of writing, this paper has been retracted from arXiv due to errors), and Chen \cite{crowdflower} in the Kaggle Crowdflower competition directly optimise the metric but in the context of gradient-boosted decision trees (XGBoost), rather than neural networks.

\subsection{Introduction}

The quadratic weighted kappa is defined as the following:
\begin{equation}
\kappa = 1 - \frac{\sum_{i,j} \textbf{W}_{ij} \textbf{O}_{ij}}{\sum_{i,j} \textbf{W}_{ij} \textbf{E}_{ij}}
\end{equation}

It measures the level of agreement between two raters, $A$ and $B$ (where one rater is the ground truth and the other is the classifier). If $\kappa = 0$ then the classifier performs no better than random chance, and if $\kappa = 1$ then there is complete agreement between the two raters. Let us now explain what each term in this equation represents.

$\textbf{O}$ is a $k \times k$ matrix where $\textbf{O}_{ij}$ is a count of how many times an instance received a rating $i$ by rater $A$ and a rating $j$ by rater $B$. This is equivalent to $\{\tilde p(A = i, B = j)\}_{i,j = 1}^{k}$, i.e., what is the (unnormalised) joint probability that rater A classifies an instance as class $i$ and rater B classifies the same instance as class $j$. If we let $\textbf{Y}$ be a $n \times k$ matrix where each row $\textbf{Y}_{i}$ denotes the one-hot encoded label of a class, and $\textbf{P} = \{p(\textbf{Y}_{i}|\textbf{X})\}_{i=1}^{n}$ be the corresponding matrix of predictions, then $\textbf{O} = \textbf{Y}^{T}\textbf{P}$.

$\textbf{E}$ is also a $k \times k$ matrix of `expected' ratings, i.e., what if we assumed that $\tilde p(A = i, B = j) = \tilde p(A = i) \times \tilde p(B = j)$? Then in that case, we can simply construct the expected ratings matrix $\textbf{E}$ (also a $k \times k$ matrix) by computing the outer product between the vector of column sums of $\textbf{Y}$ and the vector of column sums of $\textbf{P}$. let us also normalise $\textbf{E}$ so that it has the same total sum as $\textbf{O}$, by computing $\textbf{E} := \frac{\textbf{E}}{\sum_{i,j}\textbf{O}_{ij}}$.

Lastly, let us define $\textbf{W}$. This is simply a $k \times k$ matrix where $\textbf{W}_{ij}$ denotes the cost associated with misclassifying class $i$ as class $j$ (and the converse). For a discrete classification, $\textbf{W}_{ij} = 0$ for $i = j$ and 1 otherwise. For quadratic weighted kappa, $\textbf{W}_{ij} = (i-j)^2$.


Given that we have reformulated the QWK in terms of simple matrix operations on $\textbf{Y}$ and $\textbf{P}$, we can implement the fractional part of the QWK as a loss function, i.e. minimize $\frac{\sum_{i,j} \textbf{W}_{ij} \textbf{O}_{ij}}{\sum_{i,j} \textbf{W}_{ij} \textbf{E}_{ij}}$.


\subsection{Methods and Results}

\begin{figure}[H]
    \centering
    \includegraphics[width=0.9\textwidth]{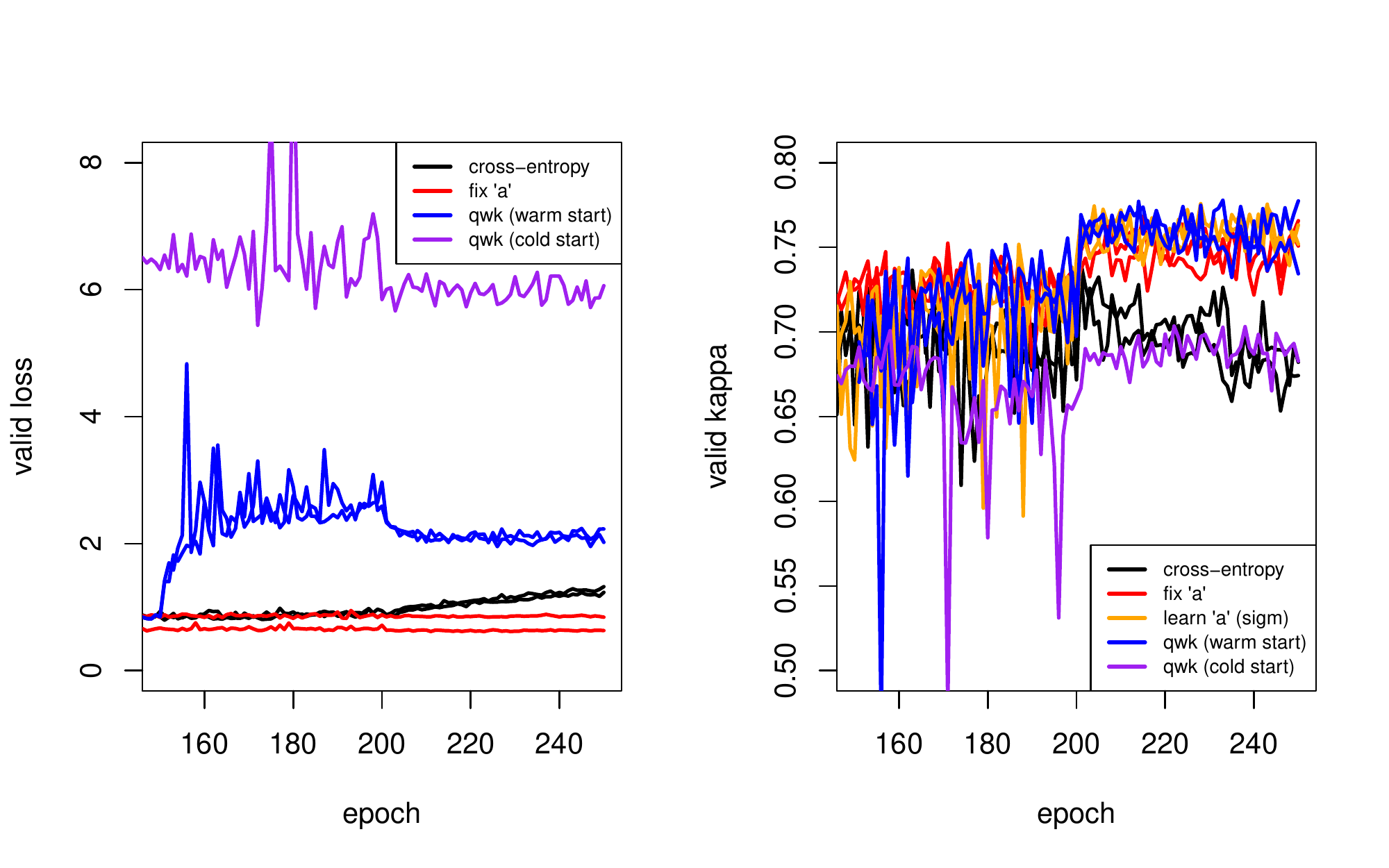}
    \caption{Cross-entropy on the validation set (left) and quadratic weighted kappa on the validation set (right) for the experiments.}
    \label{fig:qwk_exps}
\end{figure}

Figure \ref{fig:qwk_exps} shows the results of our QWK experiments. The experiment \texttt{qwk (cold start)} is when we optimise the QWK metric from start to end, and \texttt{qwk (warm start)} is when we optimise it after 150 epochs of training on cross-entropy. Just like the experiments presented earlier in this paper, for both QWK experiments we use SGD + Nesterov momentum with learning rate $\alpha = 0.01$ and switch to $\alpha = 0.001$ after 200 epochs.

Although the quadratic weighted kappa experiment produces a high cross-entropy on the validation set, the QWK on the validation set outperforms the other experiments, with the exception of \texttt{learn 'a' (sigm)}. While it can be argued that this is acceptable since cross-entropy is not the right metric to optimise (because all misclassifications have equal penalty under cross-entropy), this tells us that the QWK loss is producing rather bizarre probability distributions, since mass is being displaced from the correct classes.

To further examine the effect the QWK loss has on the cross-entropy of the validation set, we examine the probability values associated with the correct class for each example in the validation set for both the cross-entropy and QWK warm start experiment.

\begin{figure}[H]
    \centering
    \includegraphics[width=\textwidth]{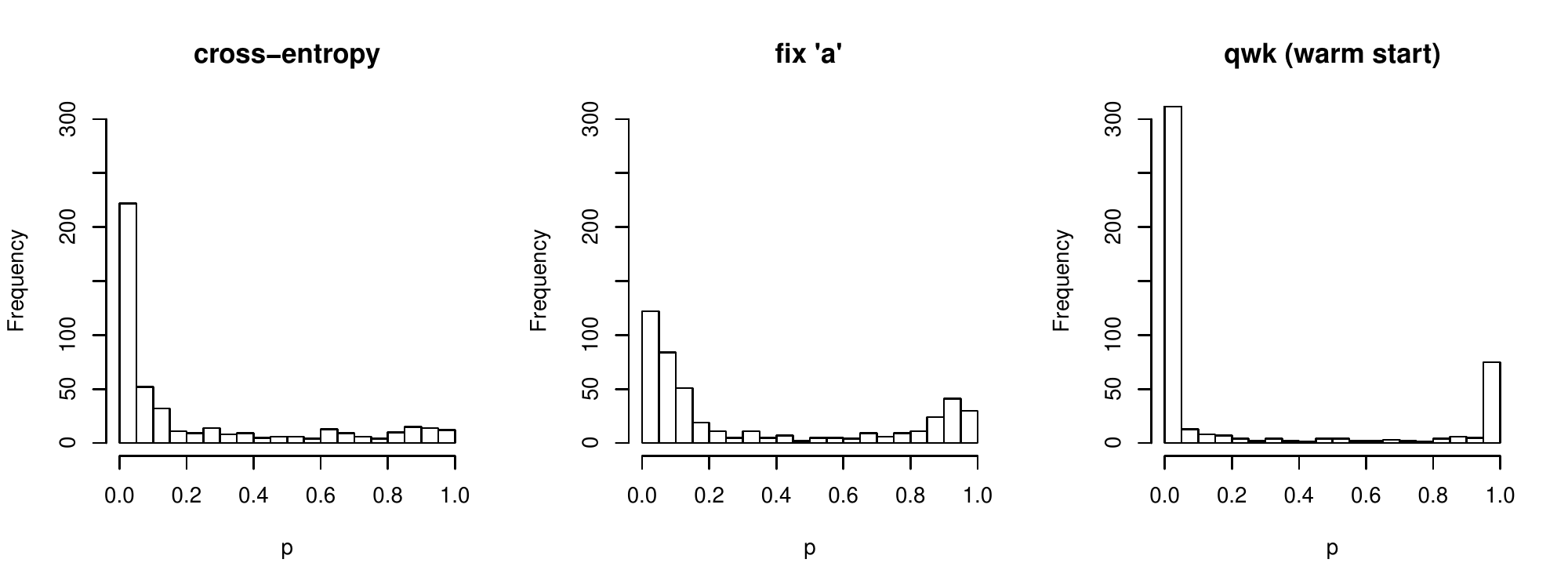}
    \caption{Histogram of probabilities for cross-entropy, our technique, and QWK (warm start), when we extract the probability associated with the correct class for each example in the validation set.}
    \label{fig:qwk_hist}
\end{figure}

Interestingly, we can see that for the QWK experiment, the number of probability estimates associated with the correct class that are close to zero are higher than the cross-entropy experiment (cross-entropy has just over 200 and QWK has just over 300). This seems to explain why the cross-entropy on the validation set is higher under the QWK experiment. While this is unfortunate, the QWK histogram also has a lot more probability values that are close to one, which may explain why it has a QWK superior to those in the other experiments. Our technique, \texttt{fix 'a'}, seems to lie somewhere in between cross-entropy and QWK: it produces the least probabilities close to zero (around 125), but produces more probabilities close to one than cross-entropy.

Lastly, we examine the distribution of predicted probabilities associated with each class under the validation set. This is shown below in Figure \ref{fig:pdist_boxplots}.

\begin{figure}[H]
    \centering
    \includegraphics[width=\textwidth]{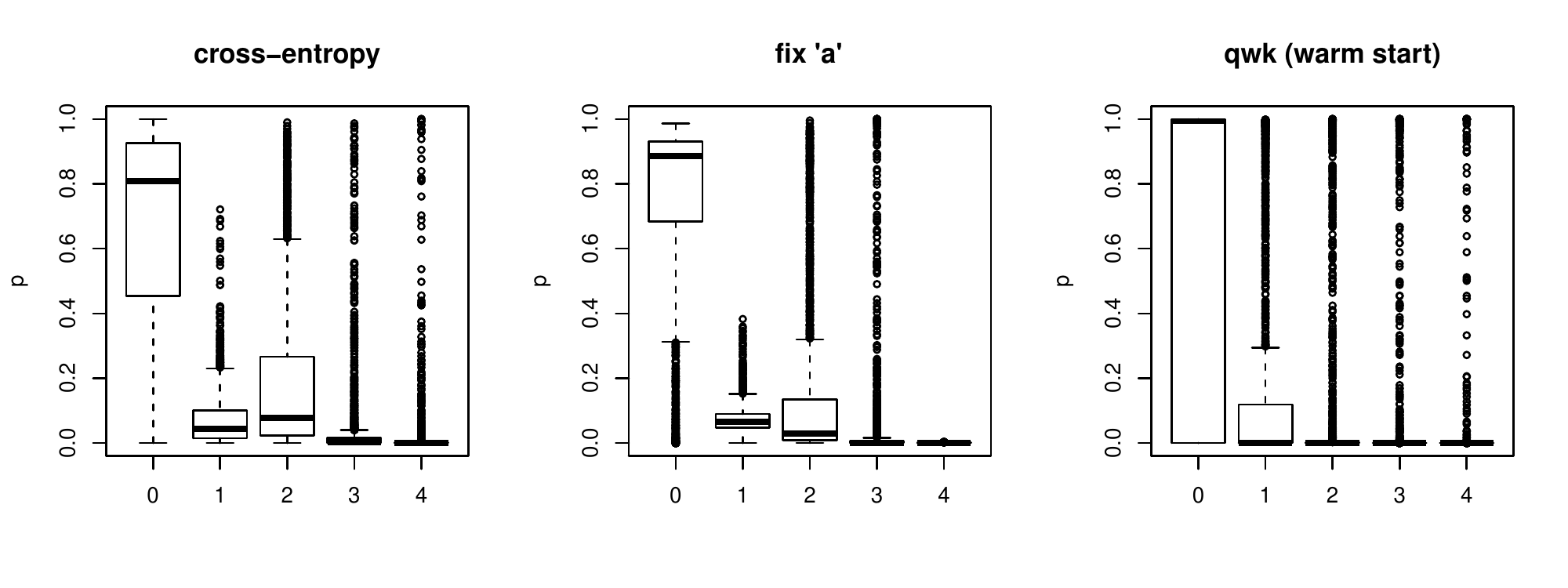}
    \caption{Boxplots depicting distributions of predicted probabilities for each class in the validation set.}
    \label{fig:pdist_boxplots}
\end{figure}

First, we compare cross-entropy to our technique, \texttt{fix 'a'}. We can see that our technique has a greater aversion to predicting high probabilities for classes 1, 2, 3, and 4. This appears to reflect the more `conservative' nature of our technique. The probabilities for \texttt{qwk (warm start)} reflect an even more conservative nature however: the median probabilities for class 1 and 2 are less than those for \texttt{fix 'a'}, and the median probability for class 0 is very close to one.

\bibliographystyle{unsrt}
\bibliography{sample}

\end{document}